# Conditional GANs for Multi-Illuminant Color Constancy: Revolution or Yet Another Approach?

Oleksii Sidorov
The Norwegian Colour and Visual Computing Laboratory, NTNU
Gjøvik, Norway
oleksiis@stud.ntnu.no

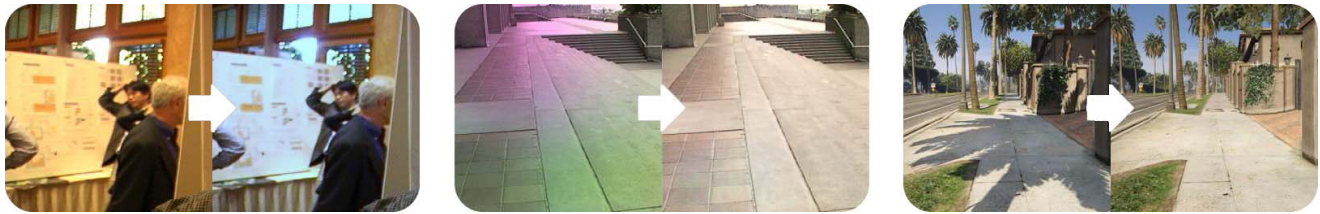

Figure 1: Examples of color constancy tasks discussed in this work. Outputs generated using the proposed algorithm.

## Abstract

*Non-uniform and multi-illuminant color constancy are important tasks, the solution of which will allow to discard information about lighting conditions in the image. Non-uniform illumination and shadows distort colors of real-world objects and mostly do not contain valuable information. Thus, many computer vision and image processing techniques would benefit from automatic discarding of this information at the pre-processing step. In this work we propose novel view on this classical problem via generative end-to-end algorithm based on image conditioned Generative Adversarial Network. We also demonstrate the potential of the given approach for joint shadow detection and removal. Forced by the lack of training data, we render the largest existing shadow removal dataset and make it publicly available. It consists of approximately 6,000 pairs of wide field of view synthetic images with and without shadows.*

## 1. Introduction

The human visual system has an inherited ability to discard information about illumination and perceive colors almost unchanged independently of the ambient conditions. This phenomenon is called color constancy (CC). The artificial algorithms of the computational color constancy aimed at correcting the effect of illumination and extraction the actual object color value in the scene as it would appear under the canonical light. However, the vast majority of computational CC algorithms work under the assumption that illumination on the scene is uniform, *i.e.* its color has the same values in each pixel of an image. This assumption is rarely fulfilled in real-world scenes. Starting from a single light source, the illuminance of which depends on the distance to the surface, up to multiple light sources in the scene – all these cases require an estimation of illumination map containing pixel-wise information. This task is highly non-trivial and is still not solved despite the long study.

Uniform CC algorithms typically provide a vector of illumination color as an output. Non-uniform CC methods do the same for smaller regions (patches or superpixels). Afterward, the colors of the scene can be corrected using diagonal transformation (dividing pixel values by illumination coordinates channel-wise). In this work, we concentrate on the statement that image with corrected colors (as they look under canonical illumination) is the main goal of any color constancy algorithm, while the estimation of illumination is an intermediate step. Thus, we propose a radically different end-to-end approach, which generates images with corrected colors directly, avoiding estimation of illumination color or illumination map. This is possible via supervised learning of a mapping between images under unknown and canonical illumination. The algorithm obtained in result demonstrates outstanding performance even in the advanced case such as random distribution of multiple illuminants, and, we believe, may be a breakthrough in this domain.

Inspired by success in discarding the complex illumination distributions we try to apply the same technique for the task of shadow removal, which can be considered a particular scenario of two-illuminant CC. However, it is impossible to remove shadows from real-world scenes completely, that causes a significant obstacle on the way of creating a large training dataset. The largest available datasets with a ground truth information are limited to a few hundreds of images, which may not be sufficient for a deep learning model. Therefore, we created a custom dataset of almost 6,000 images of real-world-like scenes rendered using advances in computer graphics. We make it publicly available and believe it will be helpful for future shadow removal research. We trained the model using this data and demonstrated that it has potential in shadow removal application as well.

Overall, our contributions are as follows:
- We are first who propose to use the generative adversarial network for solving computational color constancy task end-to-end, without an estimation of illumination color or illumination color map.
- We propose novel architecture called AngularGAN oriented specifically to CC task.
- We create and make openly accessible the largest available shadow removal dataset, and demonstrate the potential of using the described approach for shadow detection and removal.

## 2. Related works

2.1. Color Constancy under uniform illumination

The classical color constancy algorithms work under an assumption that illumination $e(x, y)$ is uniform all around the image area $e(x, y) = e$. Consequently, estimation of the illumination color vector $e = (e_R, e_G, e_B)^T$ allows to perform color correction using a diagonal model, or von Kries [46] transformation:

$$I^C(x, y) = \begin{bmatrix} e_R^{-1} & 0 & 0 \\ 0 & e_G^{-1} & 0 \\ 0 & 0 & e_B^{-1} \end{bmatrix} I(x, y), \quad (1)$$

where $I$ is the image taken under an unknown light source, $I^C$ is the image corrected, and a canonical illumination [19] is taken as $e^C = (1, 1, 1)^T$.

As the simplest case of color constancy, uniform color constancy was studied in details, and for now, may be solved with satisfactory accuracy. The studies performed in this field can be separated into two groups: statistics-based and learning-based. Methods from the first group were widely used in last decades and exploit statistics of a single image. Usually, they apply strong empirical assumptions and operate in their limits. For instance, the White Patch (WP) algorithm [8] is based on the assumption that the brightest point of an image is a perfect white reflector, Grey World (GW) algorithm [9] is based on the assumption that the average color of a scene is achromatic, Grey Edge (GE) algorithm [73] exploit the assumption that edges on the image are achromatic, *e.g.*, by applying GW algorithm to 1st or 2nd derivative of an image. All of them were generalized in one framework by van de Weijer *et al.* [73]. It is worth noting that performance of the statistics-based algorithms is different for different images, and as was shown by Gijsenij and Gevers [22] – natural image statistics can be used to cluster images by suitable CC algorithm.

The first learning-based approaches include Gamut-Mapping algorithm [14], Colour-By-Correlation [15], Exemplar-based algorithm [39], Bayesian color constancy [7], and a number of neural networks-based approaches which utilize hand-crafted features[10][11]. Further rise of deep learning algorithms has not gone unnoticed and generated a group of CNN-based color constancy algorithms. Such as: patch-based CNNs by Bianco *et al.* [4] and Shi *et al.* [65], fine-tuned AlexNet [47] by Lou *et al.* [55], custom Mixed Max-Minkowski pooling network by Fourure *et al.* [20], and current state-of-the-art $FC^4$ algorithm by Hu *et al.* [32].

2.2. Color constancy under non-uniform illumination

Uniformity of the illumination by the scene area is a very strong assumption which is usually violated in real-world scenes. Non-uniform distribution of single or multiple light sources may be generalized as one task: estimation of illumination map $e = e(x, y)$ which describes illumination color in each pixel of the image independently. A number of algorithms were proposed to solve this task, however, due to the greater complexity and ambiguity available solutions are usually limited by strict assumptions and can produce satisfactory accuracy only under particular conditions, but are not suitable for general use.

The obvious approach to non-uniform color constancy is to split the image into regions/patches/superpixels within which illumination can be considered uniform and apply conventional uniform CC algorithms to each region independently. This approach is exploited in work of Ebner [13], who assumes a grey-world assumption works locally (which is even less likely than GW assumption for a wide scene), Bleier *et al.* [5] proposed to segment an image into a set of superpixels based on color, Gijsenij *et al.* [23] proposed to obtain image patches by grid-based, keypoint-based, or segmentation-based sampling, and then estimate the illuminant for each image patch by one of the grey-based methods. Xiong *et al.* [76] using the Retinex algorithm [51] while assuming that illumination varies smoothly across the scene. In [2], Barnard *et al.* used smoothness constraints on both the reflectance and illumination gamuts to identify varying illumination. Gu *et al.* [27], on the other hand, group pixels into regions that jointly maximize the weighted sum of the illuminants in the scene and the likelihood of the associated image reflectance. A number of other studies impose additional constraints such as number of lights that lit the scene [31], capturing the near-infrared signal [67], employment of specialized hardware [16] or user inputs [6].

Recent works in this domain include those of Beigpour *et al.* [3] who use conditional random field to combine local illuminant interactions with their global spatial distribution; Mutimbu and Robles-Kelly [59] who propose a few algorithms based on factor graphs for recovering the pixelwise illuminant; and Hussain and Akbari [37] who propose an algorithm that uses the normalized average absolute difference of each segment as a measure for determining whether the segment's pixels contain reliable color information.

## 2.3. Shadow detection and removal

All the algorithms dedicated to color correction of shadowed regions can be separated in two sub-tasks: shadow detection, which produces binary shadow mask as output, and shadow removal (shadow lightning, color correction) itself. Traditional shadow detection methods [60][64][69] exploit physical models of illumination and color. However, due to the approximations in the physical model, their performance is limited. Other approaches learn shadow properties under supervision using hand-crafted features such as color [28][50][70], texture [28][70][80], edge [34][50][80], and T-junction [50]. Guo *et al.* [29] adopt similar features but detect shadows by classifying segments in an image and pairing shadow and lit segments globally, which increases the algorithm's robustness.

Recent algorithms take advantage of the representation learning ability of Convolutional neural networks (CNNs) to learn hierarchical features for shadow detection. Khan *et al.* [42] used multiple CNNs to learn features in super pixels and along object boundaries. Vicente *et al.* [71] trained stacked-CNN using a large dataset with noisy annotations. Hosseinzadeh *et al.* [30] detected shadows using a patch-level CNN and a prior shadow map generated from hand-crafted features. Qu *et al.* [61] proposed DeshadowNet with a multi-context architecture, where the output shadow matte is predicted by embedding information from global view, appearance, and semantic information. Hu *et al.* [33] use a spatial Recurrent Neural Network with attention weights. Other methods perform detection using user-hints such as clicks or strokes on the shadowed regions [24][74][78].

Removing the shadow after detection is conventionally performed either in the gradient domain [17][18][54][58] or the image intensity domain [1][24][29][41]. Guo *et al.* [29] remove shadows by image matting; Xiao *et al.* [75] apply a multi-scale adaptive illumination transfer which performs well for removing shadows cast on surfaces with strong texture; Zhang *et al.* [78] remove shadows by aligning the texture and illumination details; Khan *et al.* [41] apply a Bayesian formulation to robustly remove common shadows, however, this method is unable to process difficult shadows such as non-uniform shadows, and also computationally expensive. In work of Gong and Cosker [24], shadow removal is performed interactively by registering the penumbra to a normalized frame which allows estimation of non-uniform shadow changes.

## 2.4. Image generation using GANs

The recent development of Generative Adversarial Networks (GANs) [25] gave rise to a new era in synthetic image generation. GANs have shown remarkable results in various computer vision tasks such as image generation [35][40][62][79], image translation [38][43][81], video translation [72], deblurring [48], segmentation [56], super-resolution imaging [52], and face image synthesis [44][53]. A core principle behind any GAN model is a competition between two modules: a discriminator and a generator. The discriminator learns to distinguish between real and fake samples, while the generator learns to generate fake samples that are indistinguishable from real samples. GAN-based *conditional* image generation has also been actively studied. Pix2pix algorithm by Isola *et al.* [38] learns mapping between pairs of images from different domains in a supervised manner using cGAN [57]. The generator has a "U-Net"-shaped architecture [63] with skip-connections, while the discriminator is a convolutional classifier. Pix2pix algorithm demonstrated remarkable performance in the mapping of very different domains (labels ↔ photo, map ↔ aerial photo, edges → photo, BW → color photos, *etc.*) [38]. Motivated by these achievements, we study the possibility of using image translation for the CC task by mapping images under unknown and canonical illumination. Pix2pix learns mapping in a supervised manner, thus, require paired images for training. CycleGAN [81] and DiscoGAN [45] are examples of unsupervised algorithms which do not require paired data thanks to utilizing a cycle consistency loss. In the case where paired data is available, there is no motivation to use unsupervised algorithms and introduce additional uncertainty, however, in future it may be beneficial to collect a very large dataset of unrelated images under different illuminations and learn mapping between them.

## 3. Methodology

In this work, we present a novel architecture called AngularGAN oriented specifically to the CC task (Fig. 2). The additional criterion evaluates an illumination map from the predicted image and compares it to the ground truth during training. This technique allows to minimize angular error explicitly and was shown by Hu et al. [32] and Sidorov [66] to be efficient for the CNN-regressors. Particularly, the total loss is constructed as a linear combination of discriminator loss, L1-loss, and angular loss:

$$L = L_D + \lambda_{L1}L_1 + \lambda_{ang}L_{ang}, \qquad L_{ang} = mean(\varepsilon_{ij}) \quad (2)$$

Please note that AngularGAN generates a corrected version of the input image directly, while illumination is estimated afterwards and used only for training purposes. Such an approach allows to avoid artificial assumptions and hypotheses in regards to the illumination distribution.

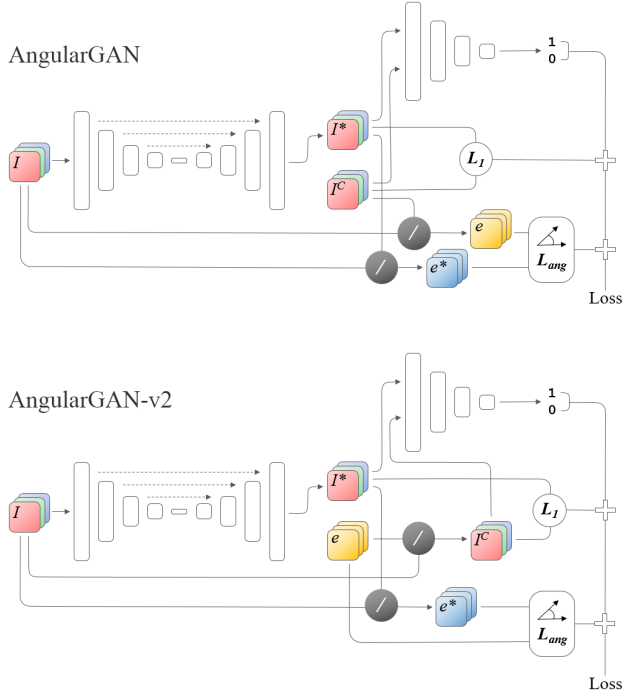

Figure 3: AngularGAN framework. $I$, $I^C$, and $I^*$ correspond to input, ground truth, and predicted images, whereas $e$ and $e^*$ are ground truth and estimated illumination maps respectively. Sign / denotes pixel-wise division.

Considering that illumination is not predicted directly, it can be estimated pixel-wise using inverse diagonal transformation and then used for calculation of the error:

$$\varepsilon_{ij} = \cos^{-1} \frac{e_{ij} \cdot e^*_{ij}}{\|e_{ij}\| \cdot \|e^*_{ij}\|}, \qquad e_{ij} = \frac{I_{ij}}{I^C_{ij}}, e^*_{ij} = \frac{I_{ij}}{I^*_{ij}} \quad (3)$$

However, this is invalid for black and over-saturated pixels which does not allow to reach zero error (Table 1). Thus, it is preferable to use ground truth illumination map when available. For such cases, we propose AngularGAN-v2 which uses illumination maps for training and implicitly computes $I^C$ as $I^C_{ij} = e_{ij}^{-1} \cdot I_{ij}$. This model produces a lower angular error but requires ground truth illumination data which may not be available in a common case.

Due to the limitation of the format, more details as well as source code can be found on the project page.[1]

## 4. Uniform Color Constancy

Firstly, the ability of a GAN to solve the classical uniform color constancy task was studied. The general idea is to learn mapping between scenes under unknown and canonical illumination, and further generate color corrected images without an intermediate step of estimation of illumination color.

---

[1] Source code and datasets: https://github.com/acecreamu/angularGAN

## 4.1. Datasets

Taking into account that accurate image generation requires a large amount of learning data we selected the largest of standard benchmark datasets – SFU Grayball [12] dataset. It contains 11,346 real-world images. In each image, a gray ball is placed in the corner of the image to provide ground truth illumination information. This information was used to discard color cast via von Kries transform (Eq. 1) and obtain color corrected images. In the preprocessing step, the data was square-cropped and the gray ball was removed.

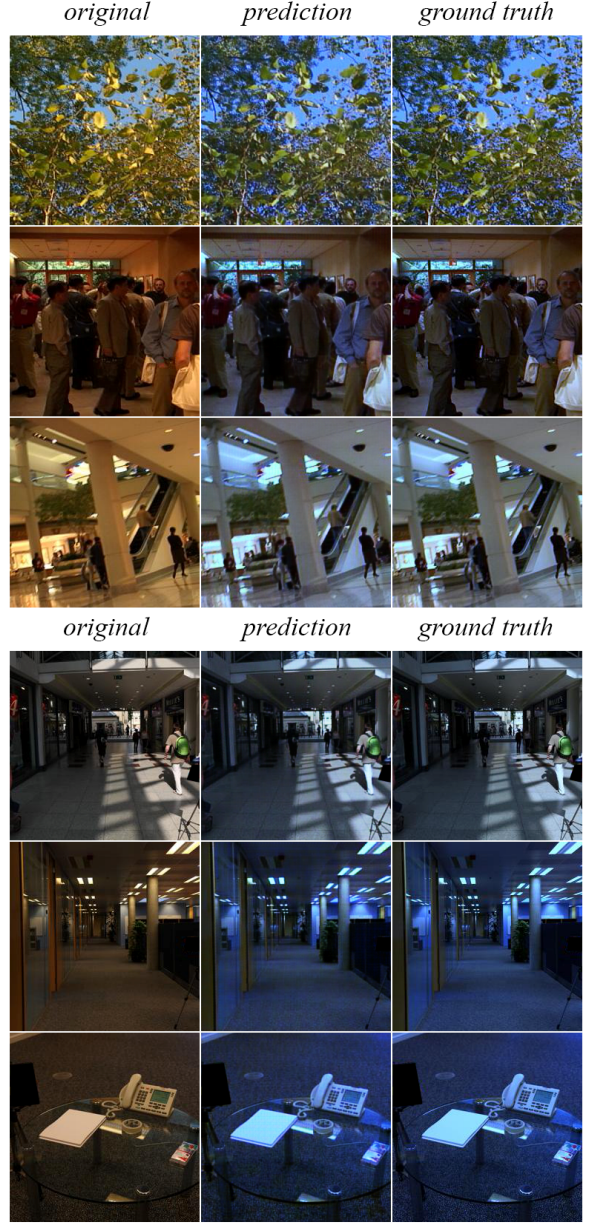

Figure 3: Results produced by the proposed approach. Top: SFU Grayball dataset (10,590 train + 756 test images); bottom: ColorChecker dataset (365 train + 183 test images).

TABLE 1. Comparison of uniform color constancy methods.

|  | SFU Grayball [12] | | ColorChecker [21] | |
| --- | --- | --- | --- | --- |
|  | mean | std | mean | std |
| Do nothing | 9.00° | 7.02° | 11.2° | 8.70° |
| GW | 8.44° | 5.35° | 8.24° | 4.51° |
| WP | 8.03° | 5.74° | 9.51° | 7.28° |
| Shades of Gray | 7.24° | 4.37° | 8.26° | 5.53° |
| GE 1st order | 8.04° | 5.74° | 9.80° | 6.74° |
| GE 2nd order | 8.19° | 6.16° | 9.93° | 7.49° |
| Weighted GE | 8.61° | 6.44° | 10.0° | 7.02° |
| Gamut Mapping | 8.15° | 5.55° | 8.81° | 6.09° |
| Exemplar Based | 7.30° | 4.86° | 8.23° | 5.45° |
| CNN | 4.36° | 2.93° | 7.18° | 3.29° |
| **AngularGAN** ($\lambda_{L1} = 0$) | 5.12° | 3.68° | 7.52° | 3.00° |
| **AngularGAN** ($\lambda_{ANG} = 0$) | 6.74° | 4.06° | 6.16° | 2.98° |
| **AngularGAN** | 4.67° | 3.08° | 6.09° | 2.59° |
| Ground Truth | 3.50° | 2.46° | 4.60° | 2.13° |

Additionally, we also apply the given approach to the much smaller ColorChecker dataset [21] which is standard in CC research. This dataset contains 548 real-world images, among which only 360 were used for training. Thus, we did not expect good performance due to the insufficient amount of learning samples. However, results demonstrate that the given approach may produce competitive accuracy even with such an extremely small learning base.

### 4.2. Results

Following previous works, we report results in terms of angular error, which is an angle between vectors of ground truth ($e$) and estimated ($e^*$) illumination color. However, we doubt the adequacy of this metric for the algorithms like ours, where $e^*$ can be estimated only in approximation, and encourage the community to use image similarity metrics such as SSIM, PSNR, and CIE $\Delta E$. We compare the results with classical single-image and learning-based algorithms. As an instance of a typical CNN-based algorithm, we implement custom regressor using fine-tuned GoogLeNet [68] (by analogy with fine-tuning of AlexNet by authors of [4] and [55]). Statistical values of error metric for both datasets are presented in Table 1, sample images for visual evaluation are shown in Figure 3.

The quantitative evaluation shows that our approach produces results competitive to classical CC algorithms and outperforms most of them. It can also be seen from the experiment with ColorChecker dataset that requirements for the size of the training set are not as strict as it may be expected for a generative algorithm.

It should be noted that due to the generation of output images instead of simple modification, given approach may introduce loss of image quality that is not typical for conventional CC algorithms. This may manifest as artifacts and periodic noise. Although they are not visible with the naked eye and expectedly will be solved in next generations of GANs, this peculiarity of image generation should be taken into account.

## 5. Multi-Illuminant Color Constancy

Multi-Illuminant and non-uniform CC are much more complex problems in comparison to uniform CC because they require estimation of a map of illumination for each pixel of the image instead of applying one value to all of them. The proposed image-to-image translation approach to CC does not estimate the illumination map explicitly and performs color correction by learning mapping based on data provided. Thus, it is not influenced by the complexity of the illumination distribution directly. This feature allows achieving outstanding results in removing complex color cast from an image.

### 5.1. Datasets

The main multi-illuminant datasets are: Multiple Light Sources Dataset [23] (59 laboratory + 9 outdoor images), Multiple-Illuminant Multi-Object dataset [3] (10 laboratory scenes under 6 conditions + 20 real-world images), and Multi-Illuminant Dataset [5] (4 laboratory scenes under 17 illumination conditions). There is also a number of smaller datasets captured in laboratory conditions. None of the available datasets is larger than 100 images, which does not allow using them for the training of deep learning-based models. Therefore, we decided to synthesize a custom dataset of an appropriate size. The color corrected images from SFU Grayball dataset [12] were taken as a ground truth data. Tint maps were created as Gaussian distributions of various (for generalization) colors with random μ and σ (Fig. 4, top row). Distorted images were created by tinting ground truth data with tint maps using inverse von Kries transform (Fig. 4, second row). In result, we obtained 11,346 images synthetically color-casted with random combinations of the three different illuminants.

TABLE 2. Comparison of accuracy of multi-illuminant color constancy methods.

|  | Angular Error | | PSNR | |
| --- | --- | --- | --- | --- |
|  | mean | std | mean | std |
| Do nothing | 9.58° | 3.41° | 16.9 | 2.55 |
| LSAC [13] | 15.2° | 4.84° | 14.3 | 2.52 |
| Gijsenij et al. [23] WP | 12.1° | 5.18° | 15.4 | 2.77 |
| Gijsenij et al. [23] GW | 10.5° | 5.00° | 17.0 | 2.61 |
| MIRF [3] | 9.32° | 3.43° | 20.5 | 2.50 |
| MICC [49] | 8.15° | 4.80° | 19.8 | 2.48 |
| Mutimbu and Robles-Kelly [59] | 6.64° | 3.90° | 22.1 | 2.31 |
| Hussain and Akbari [37] | 6.15° | 3.26° | 22.6 | 2.35 |
| **AngularGAN** ($\lambda_{L1} = 0$) | 4.75° | 4.11° | 21.6 | 2.45 |
| **AngularGAN** ($\lambda_{ANG} = 0$) | 6.26° | 2.01° | 27.5 | 2.31 |
| **AngularGAN** | 3.98° | 2.16° | 29.1 | 2.12 |

The similar real-world scenes usually have similar illumination conditions (*e.g.* photos captured in office have one type of light sources) that simplifies the training of the algorithms and allows to apply exact transformation learned from the training data to the test data. The proposed custom dataset does not have this feature because the distribution of synthetic illuminants is random in each image and do not correlate between coherent scenes. This makes the learning process even more complex. However, it does not influence single-image methods.

5.2. Results

The results are reported as mean angular error between all the pixels by analogy with the previous experiment. Performance is quantitatively compared to the performance of state-of-the-art methods and is reported in Table 2. Samples of generated images are demonstrated for visual evaluation (Fig. 4). It may be seen that the proposed technique outperforms all existing multi-illuminant algorithms. Moreover, the algorithm successfully learns mapping between domains even though input color casts were not coherent. In simple words, we can explain it as learning not just mapping between given pairs of images, but learning how to generate a correctly illuminated image by a given input with an arbitrary combination of light sources. We consider this to be the dominant feature of the proposed method. Drawbacks of the given approach include but are not limited to: demand for a big volume of training data and imperfect image quality (as discussed before). Also, we observed that many images produced had slightly increased brightness, although the color cast is removed correctly. This effect influences reported difference metrics, however visually only the lightness of colors does not match, but not the hue.

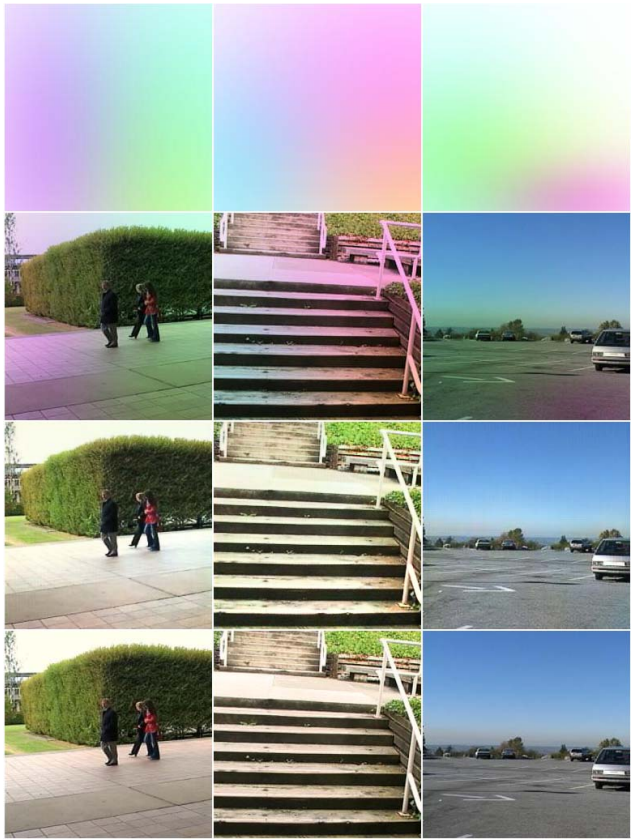

Figure 4: Results produced by the proposed approach on removing of multi-illuminant color cast. Top row – tint maps; second row – input images; third row – predictions; bottom row – ground truth.

**6. Shadow removal**

Shadow removal may be considered a special scenario of multi-illuminant color constancy. The direct light beams can be considered as a first illuminant, scattered beams of light – as a second, while the canonical illumination is taken to be equal to the first one. The chromatic distribution of illumination is trivial and of no essential interest. However, the spatial distribution of shadows may create a complex map which is defined by relief and 3D shape of the objects. Estimation of shadow map (shadow detection) therefore presents an independent complex task which may be followed by a third-party algorithm for shadow removal (color correction). By analogy with multi-illuminant CC, the approach proposed in this work allows to avoid these procedures and learn mapping between scenes with and without shadows based only on paired data samples, and perform shadow removal end-to-end.

6.1. Dataset

The standard real-world datasets for shadow removal are very limited in their size: UCF shadow dataset [82] (245 images), SRD VSC [24] (214 images), UIUC [29] (76 images), LRSS [26] (37 images). Mainly, it is caused by the difficulty of capturing the same real-world scene with and without shadow. Moreover, it is impossible to remove shadows from wide field of view scenes such as street view. This causes a specific appearance of shadow removal data similar to cropped patches (Fig. 5). There are also attempts to create large shadow removal datasets. SBU Shadow dataset [71], for example, contains 4089 real-world images, however without a ground truth; authors propose shadow mask detected using their algorithm, which then can be used for color correction by a third-party algorithm; resulting shadow-free images cannot be considered a ground truth, but only an output of two artificial algorithms, and training the model on such data in the best case will allow to achieve the performance of the used algorithms, but will not outperform it. SRD dataset by Qu *et al*. [61] is claimed to contain 3,088 shadow and shadow-free image pairs; however, two years after publication only a test set of 408 images is publicly available, which makes full use of this dataset impossible.

Eventually, forced by a significant lack of data we created a synthetic dataset of 5,723 image pairs which make it the largest shadow removal dataset available. We used computer graphics from a video-game GTA V by Rockstar to render real-world-like scenes in two editions: with and without shadows (Fig. 6). The proposed approach accurately models real world and allows to obtain fair shadow-free data for the general scenes which is impossible to implement in real life. Moreover, it captures scenes in a conventional wide field of view and avoids using small areas and patch-like appearance. Generated samples are 8bit RGB images with a 600x800 pixels resolution. The dataset contains 5,110 standard daylight scenes and additional 613 indoor and night scenes.

We also used 408 real-world images from SRD dataset for training in order to estimate model's demands to the size of the training set.

### 6.2. Results

Results are reported in PSNR and angular error between predicted and ground truth shadow-free images. The accuracy is compared with state-of-the-art methods the codes of which are available publicly. It is noteworthy that despite similar values of error metrics provided, the nature of mistakes and errors produced by algorithms is totally different. For instance, single-image methods produce errors mainly due to wrong shadow map detection, while the proposed end-to-end approach identifies shadows correctly but may generate data of low image quality.

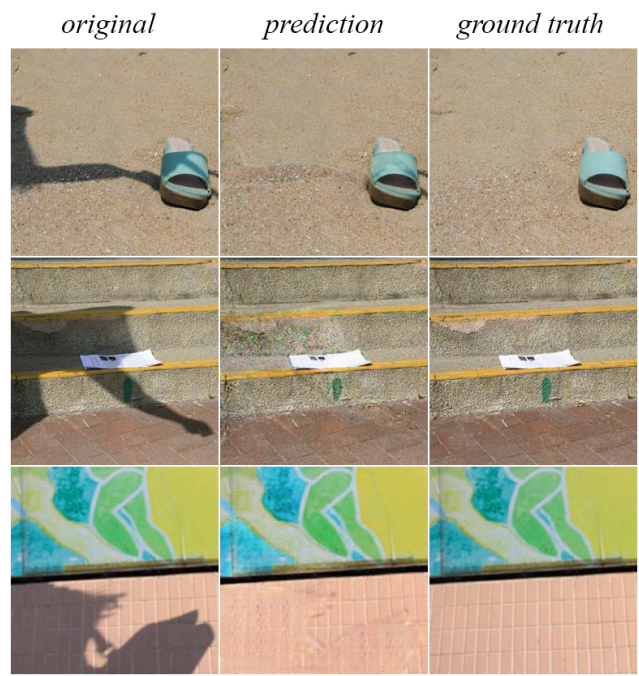

Figure 5: Results of shadow removal from real-world images. Demonstrated on SRD-test dataset (388 train and 20 test images). Quality of the output is considered to be unsatisfactory.

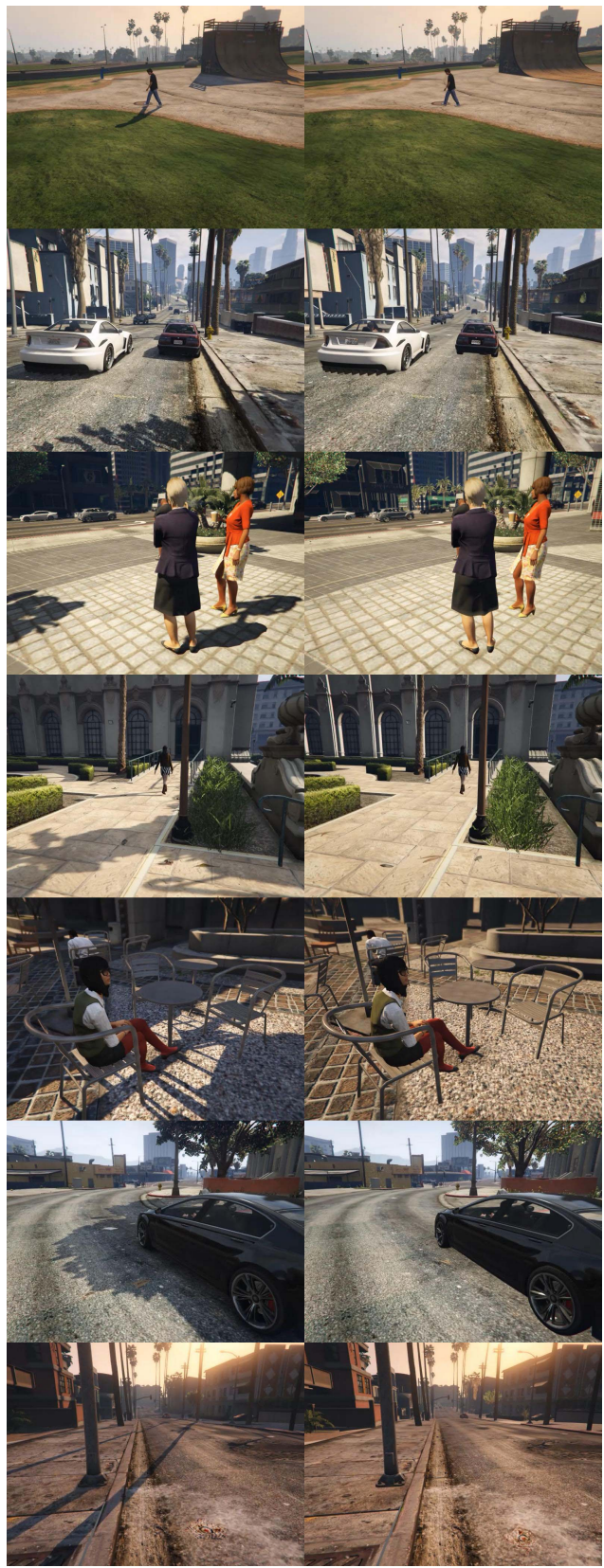

Figure 6: Preview of the proposed GTAV dataset.

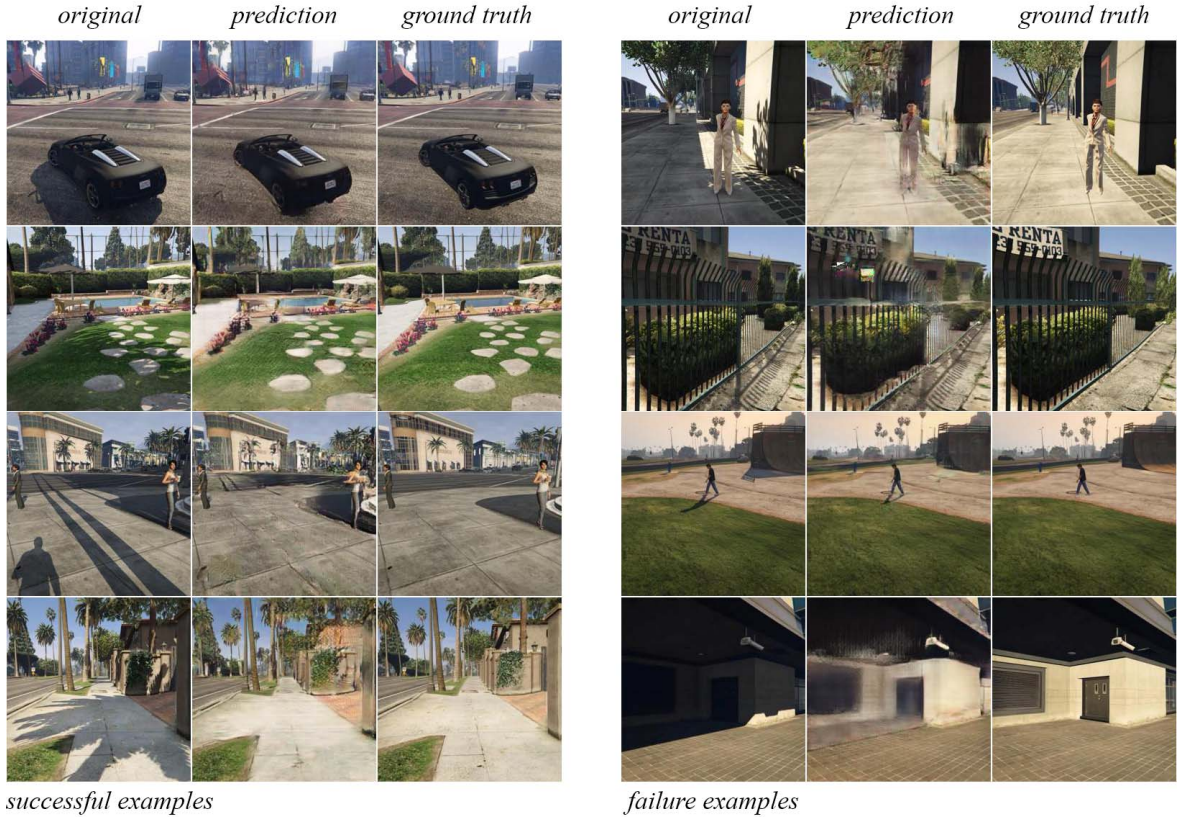

*successful examples*            *failure examples*

Figure 7: Results of shadow removal from synthetic scenes. Demonstrated on GTAV dataset (4610 training + 250 test images).

The model trained on GTAV dataset demonstrates a satisfactory accuracy of shadow removal with examples of both successful and failure cases (Fig. 7). In the worst cases, pix2pix generates artifacts and distorts original image significantly. Particular problems were observed in the cases of: ambiguous dark colors; utterly dark areas where color information is almost lost; reconstruction of high spatial frequencies. Uniform shadows of a large area approximate the case of uniform CC, and typically demonstrate better reconstruction quality.

The quantitative evaluation of results produced by the model trained on 388 and tested on 20 RSD-test images is not reported because even visually (Fig. 5) the quality of the produced output is not as good as the output of corresponding single-image algorithms. This demonstrates an increase in model's demands for the size of a training set in comparison to the simpler task of uniform CC.

## 7. Conclusions

In this work, we propose a novel, end-to-end approach to advanced cases of computational color constancy. The proposed technique is driven by data and avoids utilizing any artificial assumptions or hypotheses. It is shown to be efficient in the removal of complex illumination color cast created by multiple non-uniformly distributed light sources.

The accuracy of predicting a single "uniformly distributed" illuminant is on a par with existing methods. We explain it by over-simplification of the task that created accurate algorithms which work very good within the model, but not outside of it. Our algorithm treats both tasks equally, which make it dominant in more realistic scenarios. For example, the same algorithm can even be applied to the removal of complex shadow cast, however, the limitation of the training data is an essential obstacle.

Further development may include collecting/synthesis of larger datasets, evaluation of models trained using synthetic data on similar real-world images, and, of course, designing new generative algorithms to improve image quality.

TABLE 3. Comparison of accuracy of shadow removal methods on GTAV dataset.

|  | Angular Error | | PSNR | |
|---|---|---|---|---|
|  | mean | std | mean | std |
| Do nothing | 4.01° | 3.15° | 18.2 | 4.53 |
| Yang et al. [77] | 5.20° | 2.89° | 16.4 | 4.72 |
| Guo et al. [29] | 2.95° | 2.20° | 20.7 | 3.98 |
| Gong et al. [24] (user assisted) | 2.33° | 1.48° | 24.1 | 3.10 |
| **AngularGAN** ($\lambda_{L1} = 0$) | 3.05° | 2.20° | 19.3 | 4.05 |
| **AngularGAN** ($\lambda_{ANG} = 0$) | 3.12° | 1.45° | 21.6 | 3.36 |
| **AngularGAN** | 2.86° | 1.58° | 21.0 | 3.43 |